\documentclass[lettersize,journal]{IEEEtran}
\usepackage{amsmath,amsfonts}
\usepackage{algorithmic}
\usepackage{algorithm}
\usepackage{array}
\usepackage[caption=false,font=normalsize,labelfont=sf,textfont=sf]{subfig}
\usepackage{textcomp}
\usepackage{stfloats}
\usepackage{url}
\usepackage{verbatim}
\usepackage{graphicx}
\usepackage{cite}
\usepackage{booktabs, multirow}
\usepackage{bm}
\usepackage{xcolor}
\newcommand\blfootnote[1]{%
  \begingroup
  \renewcommand\thefootnote{}\footnote{#1}%
  \addtocounter{footnote}{-1}%
  \endgroup
}

\begin{document}

\title{ASR Error Correction using Large Language Models}

\author{Rao Ma$^\ast$, Mengjie Qian$^\ast$, Mark Gales,~\IEEEmembership{Fellow,~IEEE}, Kate Knill,~\IEEEmembership{Senior Member,~IEEE}
}

\markboth{Journal of \LaTeX\ Class Files,~Vol.~xx, No.~xx, xxxx~xx}%
{Shell \MakeLowercase{\textit{et al.}}: A Sample Article Using IEEEtran.cls for IEEE Journals}

\IEEEpubid{0000--0000/00\$00.00~\copyright~2021 IEEE}

\maketitle

\begin{abstract}

Error correction (EC) models play a crucial role in refining Automatic Speech Recognition (ASR) transcriptions, enhancing the readability and quality of transcriptions. Without requiring access to the underlying code or model weights, EC can improve performance and provide domain adaptation for black-box ASR systems. This work investigates the use of large language models (LLMs) for error correction across diverse scenarios. 
1-best ASR hypotheses are commonly used as the input to EC models. We propose building high-performance EC models using ASR N-best lists which should provide more contextual information for the correction process. 
Additionally, the generation process of a standard EC model is unrestricted in the sense that any output sequence can be generated. For some scenarios, such as unseen domains, this flexibility may impact performance. To address this, we introduce a constrained decoding approach based on the N-best list or an ASR lattice. 
Finally, most EC models are trained for a specific ASR system requiring retraining whenever the underlying ASR system is changed. This paper explores the ability of EC models to operate on the output of different ASR systems. This concept is further extended to zero-shot error correction using LLMs, such as ChatGPT. Experiments on three standard datasets demonstrate the efficacy of our proposed methods for both Transducer and attention-based encoder-decoder ASR systems. In addition, the proposed method can serve as an effective method for model ensembling.
\blfootnote{$^\ast$ Equal Contribution.}

\end{abstract}


\section{Introduction}

Automatic speech recognition (ASR) aims to transcribe speech audio into text and is the key component for human-computer interaction~\cite{rebman2003speech}. In recent years, the performance of ASR technology has dramatically advanced, evolving from traditional Hidden Markov Model (HMM)-based architectures to modern end-to-end (E2E) systems like Listen, Attend and Spell (LAS) or RNN-T~\cite{chan2016listen, graves2014towards, amodei2016deep,vaswani2017attention}. 
Large-scale models such as Whisper~\cite{radford2023robust} and Google USM~\cite{zhang2023google} have demonstrated state-of-the-art performance, leveraging vast amounts of labeled and unlabeled speech data, which can be costly to obtain. While achieving impressive results on test sets, practical deployment of ASR systems encounters challenges, especially when faced with domain-specific or previously unseen speech data.

Accessing ASR services via APIs has emerged as a popular alternative to training in-house models, offering a pragmatic and economical choice. Fine-tuning these models for specific tasks, however, is often impractical due to restricted access to proprietary models. Various approaches have been proposed to enhance such restricted access ASR systems. Two common methods are language model (LM) rescoring and error correction (EC). LM rescoring involves reranking the N-best list generated by the ASR system using an external LM, which can improve the overall performance of the ASR system~\cite{mikolov2010recurrent}. Recent research, however, has shown that E2E ASR models often learn an internal language model (ILM) on the training data, which can reduce the effectiveness of traditional shallow fusion techniques~\cite{meng2021internal}. 
Methods to address the impact of ILMs, such as those proposed by~\cite{mcdermott2019density, liu22j_interspeech, zeineldeen2021investigating}, generally involve code modification during the inference stage, therefore they are out of the scope of discussion in this paper.



\IEEEpubidadjcol
Error correction, applied as a post-processing step for ASR systems, offers a promising alternative~\cite{errattahi2018automatic, hrinchuk2020correction, zhao2021bart}. This approach requires only the decoding hypotheses and reference data to train a model, eliminating the need for deep access to the ASR system.
Early work in this area focused on rule-based systems which rely on statistical analysis~\cite{cucu2013statistical}. More recent developments have introduced end-to-end models with attention modules, which can automatically identify errors within sentences and learn to generate the correct counterparts implicitly~\cite{ren2019fastspeech, guo2019spelling, mani2020asr}. Large-scale pre-trained language models (PLMs) are trained on massive and diverse text datasets, far exceeding the scale of data used in ASR training.
Approaches to transfer knowledge from PLMs for accurate ASR error detection and correction have been recently proposed. For example, Hrinchuk et al.~\cite{hrinchuk2020correction} propose a Transformer-based architecture to ``translate'' an ASR model output into grammatically
and semantically correct text. Zhao et al.~\cite{zhao2021bart} introduce a BART-based semantic correction system for the Mandarin ASR system. Shen et al.~\cite{shen2022mask} propose a masking strategy to train the model to correct the original error tokens and predict the masked tokens based on their context information. Ma et. al~\cite{ma2023nbest, ma2023adapting} propose an N-best T5 model based on pre-trained T5 models to perform error correction using the ASR N-best list.
By fine-tuning these pre-trained large language models (LLMs), the implicit knowledge acquired from vast amounts of text data can be effectively transferred to the target error correction task. 

The recent advent of generative LLMs has further advanced EC techniques. Within the field of NLP, studies such as~\cite{wu2023chatgpt, fang2023chatgpt} have applied ChatGPT models to grammatical error correction tasks. In the context of ASR error correction, previous research has examined zero-shot performance using LLMs~\cite{ma2023can}. Everson et al.~\cite{everson2024towards} utilized word confusion networks generated by the ASR system and performed EC with in-context learning, demonstrating improved performance with one-shot examples compared to 1-best hypotheses. Chen et al.~\cite{chen2024hyporadise} generated N-best lists in the ASR decoding and built LLM EC systems using various methods including fine-tuning, LoRA tuning, and in-context learning. 
Hu et al. developed a multi-modal EC model incorporating audio as an additional input~\cite{hu2024listen} and used a cloze-test task approach instead of a generative correction method. Additionally, Li et al. \cite{li2024investigating} explored knowledge transfer within LLMs by fine-tuning a multilingual LLM across various languages to correct 1-best hypothesis errors from different speech foundation models.

Previous research has also explored various methods to improve ASR error correction by leveraging N-best lists, which offer richer information compared to single 1-best hypotheses. For instance, Guo et al.~\cite{guo2019spelling} generates an 8-best list with the ASR model and rescored candidates with an LSTM language model~\cite{sundermeyer2012lstm}. Zhu et al.~ \cite{zhu2021improving} concatenated N-best hypotheses for input to a bidirectional encoder, and Leng et al.\cite{leng2021fastcorrect} investigated non-autoregressive models with similar approaches. More recent work by Ma et al.\cite{ma2023can} and Chen et al.\cite{chen2024hyporadise} has integrated N-best lists with generative LLMs to enhance error correction performance. 

Building on these advances, our paper introduces a novel approach that uses LLMs to improve ASR error correction. We compare fine-tuning versus zero-shot error correction methods and investigate how ASR N-best lists can be effectively utilized. A major contribution of our work is the innovative use of ASR N-best lists as extended inputs, which provides richer context and more accurate cues for error correction. Additionally, we introduce advanced decoding strategies, including constrained decoding, to enhance model robustness and alignment with the original utterance. Our approach also addresses data contamination concerns by developing methods to evaluate the impact of training data biases. By integrating these elements, our study sets a new benchmark in ASR error correction, advancing the state-of-the-art in the field.

This paper is structured as follows. In Section~\ref{sec:ec_pre} we present the error correction method utilizing foundation language models, including a supervised approach employing the T5 model and a zero-shot approach based on generative LLMs. Section \ref{sec:uncon_cons_decoding} introduces several decoding algorithms to build more robust ASR error correction systems, aiming to address the inherent problems of the standard beam search approach.  In Section~\ref{sec:contamination}, we describe the method used to investigate data contamination.
The experimental setup and results are detailed in Section~\ref{sec:exp}, while Section~\ref{sec:discussion} covers N-best analysis, an ablation study of the proposed approach, and a discussion on data contamination. Finally, Section~\ref{sec:conclusion} presents the conclusions. 

\section{ASR Error Correction Models}
\label{sec:ec_pre}

\subsection{ASR Error Correction using N-best lists}

Building upon previous findings, we explore new methodologies for effectively incorporating supplementary ASR outputs into LLM-based EC models. One promising approach involves utilizing N-best ASR hypotheses, which are generated by the ASR system as a byproduct of the beam search process~\cite{liu2021asr, ganesan2021n}. 
The integration of N-best T5, using the N-best list as input, has demonstrated significant performance improvements in error correction compared to the original model~\cite{ma2023nbest}. The rationale behind this is that the N-best list contains alternative sequences that have a strong possibility of being the correct transcription, thus providing valuable cues for the EC model during predictions. 
We modify the input of the EC model to be sentences in the N-best list, sorted based on ASR scores and concatenated with a special token, improving both interpretability and effectiveness.

In our study, we introduce methods for developing both fine-tuning and zero-shot error correction models using ASR N-best lists and propose several decoding methods. In traditional ASR error correction models, the decoding process allows the model to generate any sequence based on the given context, meaning that the output space is not constrained, which is denoted as unconstrained decoding (\textbf{uncon}).
When an N-best list is used, we can constrain the model to allow it to generate from a limited space. For this, we propose N-best constrained decoding (\textbf{constr}) and N-best closest (\textbf{closest}). The details of these decoding methods will be introduced in Section~\ref{sec:uncon_cons_decoding}.

\subsection{Fine-tuning based Approach}
\begin{figure}[!htbp]
    \centering
    \includegraphics[width=0.98\linewidth]{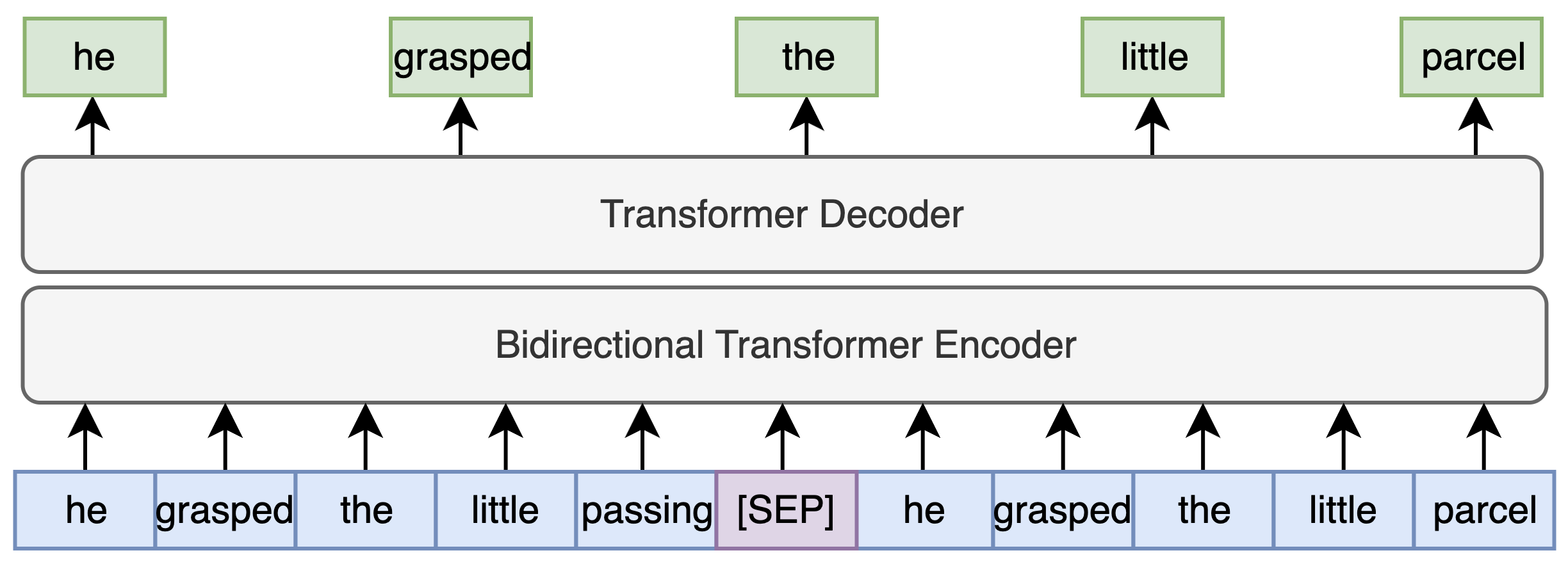}
    \caption{The model structure of a supervised error correction model using ASR N-best lists as input. Here, we set $N$ to 2 for illustration.}
    \label{fig:correction}
\end{figure}
EC models aim to correct recognition errors in ASR transcriptions, serving as a post-processing step for speech recognition.  A standard supervised EC model adopts an E2E structure, taking the ASR transcription as input, and is trained to generate the corrected sentence~\cite{guo2019spelling, hrinchuk2020correction, zhao2021bart}. 
Adapting an EC model from a PLM yields superior performance when compared to training the EC model from scratch, as this approach leverages the prior knowledge embedded in the language models~\cite{ma2023nbest}. When training an EC model, direct access to the ASR system is unnecessary as only the decoded hypotheses are required. This flexibility in data accessibility makes the method highly practical, especially in situations where adapting a black-box, cloud-based speech-to-text system is essential. 

The structure of our proposed EC model is illustrated in Figure~\ref{fig:correction}, where ASR N-best lists are given as input to the encoder and the model is trained to generate the manual reference. Here, sentences are concatenated with a special token \texttt{[SEP]}. The model is trained to automatically detect errors and generate the corrected hypothesis on specific training data. For supervised models, generalization capability is crucial, as it enhances their applicability across various practical scenarios. A model that generalizes well can be used in diverse and practical contexts without the need for further updates. In our experiments, we applied a model trained on transcriptions of corpus $A$ generated by a specific ASR system to out-of-domain test sets and outputs from other ASR systems. By doing this, we demonstrate the generalization ability of our proposed method, highlighting its robustness and adaptability across varying ASR outputs and domains.

\subsection{Zero-shot Approach}
\label{sec:ec_zero-shot}

\begin{figure*}[!htbp]
    \centering
    \includegraphics[width=0.9\linewidth]{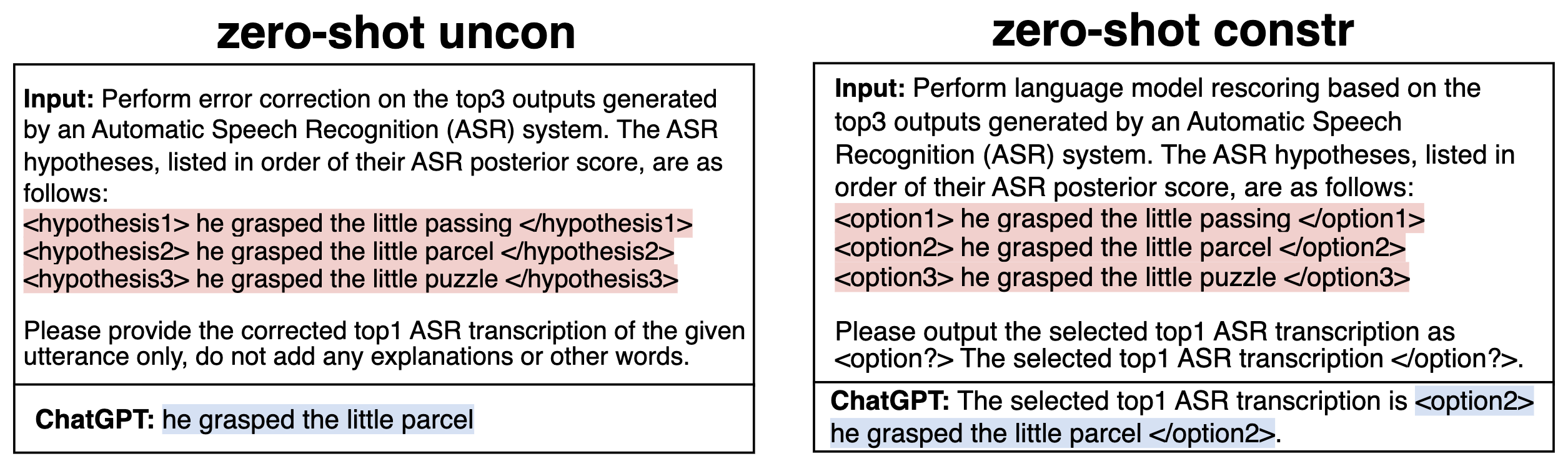}
    \caption{Prompt design for zero-shot ASR error correction. Here we use a 3-best list generated by the ASR system as input to ChatGPT for illustration.}
    \label{fig:prompt}
\end{figure*}

Supervised training has long been popular for developing EC systems, however, it requires the availability of training data and the systems can be computationally expensive to build. To address these constraints, we present our approach to utilizing generative LLMs for zero-shot error correction, using ChatGPT as an illustrative example. This task can be challenging as ChatGPT lacks prior knowledge about the error patterns of the ASR system and has no access to the original utterance. To mitigate the difficulty, similar to the supervised approach, we provide hypotheses from the N-best list as valuable hints, helping the model to detect and correct errors effectively.
The prompts we used in the experiments are shown in Figure \ref{fig:prompt}. 
In the prompt design, hypotheses are sorted by the descending ASR posterior score. Additionally, tags such as \texttt{$<$hypothesis1$>$} and \texttt{$<$/hypothesis1$>$} enclose each N-best hypothesis. We experimented with other input formats, such as using numbers instead of tags or employing plain sentences without explicitly specified order, but these variants showed degraded performance compared to our chosen prompt.
In the ablation study detailed in Section~\ref{sec:ablation_n}, we highlight the importance of using a reasonable number of N for the model to achieve optimal performance. When only the top one ASR hypothesis is used as input, ChatGPT-based error correction may experience a degradation in performance.
Additionally, initial experiments explored the few-shot setting, revealing unstable performance improvements compared to the zero-shot approach and higher computational costs. Therefore we mainly focus on the zero-shot results in this paper. 


\section{ASR Error Correction Decoding}
\label{sec:uncon_cons_decoding}
Previous sections introduced the proposed fine-tuning and zero-shot error correction methods. Specifically, we highlighted how to incorporate more information into the input space of the proposed model with ASR N-best lists and evaluate the impact on performance. Another intriguing aspect is guiding the decoding process to achieve controllable generation. In this section, we delve into this challenge, exploring methods to direct the model's output in a more controlled and predictable manner.

\subsection{Unconstrained Decoding}
For an ASR error correction model with parameters $\bm \theta_\text{EC}$, the N-best input is denoted as $\bm{\mathcal{Z}} = \{\bm{\hat z}^{(1)}, \bm{\hat z}^{(2)}, \cdots, \bm{\hat z}^{(n)}\}$. Our decoding objective is to find $\hat{\bm y}_\text{uncon}$ that satisfies
\begin{equation}
\begin{split}
     \hat{\bm y}_\text{uncon} &= {\arg\max}_{\bm y}\log P(\bm y|\bm{\mathcal{Z}};\bm \theta_\text{EC})
\end{split}
\end{equation}
where $\bm y$ presents potential output sequences. Given the computational cost of finding the globally optimal sequence, heuristic algorithms such as beam search are commonly used for decoding. In this context, the decoding method is referred to as unconstrained decoding (\textbf{uncon}), as no explicit constraints are applied to the generated sequences.



Beam search is a practical tool for approximating optimal generation results across the entire decoding space, balancing efficiency and performance. However, this method grants the model too much freedom, limiting our control over the decoding process \cite{dathathriplug}. Specifically, we aim for the proposed model to retain correct words from the original transcription and only correct inaccurate ones. In addition, the model is expected to generate homophones for detected erroneous words. While these aspects can be implicitly learned from the training data, there is no guarantee they will be applied during decoding. The model might produce synonyms with high embedding similarity to words in the reference text, which can be problematic. 
To address these concerns, we introduce several alternative decoding algorithms to enhance the model's ability to achieve the desired decoding objectives. These methods aim to exert more control over the output, ensuring that corrections are precise and that the generated sequences meet specific criteria for accuracy and relevance.

\subsection{N-best Constrained Decoding}

In unconstrained decoding, the decoding space of the EC model is unbounded. However, we want the generated correction results to closely 
resemble the original utterance. One method to introduce constraints on the decoding space involves leveraging the ASR N-best list, which comprises the top N hypotheses generated by the ASR system, representing the transcriptions most likely to be correct given the input audio. This approach, denoted as N-best constrained decoding (\textbf{constr}), forces the model to only generate sentences within the ASR N-best list.

Furthermore, each path in the N-best list is associated with a score calculated by the ASR system, indicating the likelihood of it being the correct output. These scores can be combined with the scores from the EC model, using an interpolation weight $\lambda$ to gain insights from both models. To be more specific, the decoding result $\hat{\bm y}_\text{constr}$ is derived by maximizing the equation:
\begin{equation}
\begin{split}
    \hat{\bm y}_\text{constr} &= {\arg\max}_{\bm y\in \bm{\mathcal{Z}}} [(1-\lambda) \cdot \log P(\bm y|\bm{x};\bm\theta_\text{ASR})\\
    &+ \lambda \cdot \log P(\bm y|\bm{\mathcal{Z}};\bm \theta_\text{EC})]
\end{split}
\label{eq:n-best}
\end{equation}
where  $\bm x$ and $\bm{\mathcal{Z}}$ denote the input acoustic features of the ASR system and the obtained ASR N-best list, respectively. When $\lambda$ is set to $\bm 1$, the scores from the ASR system are ignored, and only the probabilities from the EC model are considered.
This approach requires obtaining the probability scores from the EC model, which can be implemented in the supervised EC method. Section~\ref{sec:exp-t5} will demonstrate its effectiveness and highlight situations where its utility becomes evident. 

In zero-shot EC scenarios, the method is applied differently. Instead of generating a correction from scratch, ChatGPT is tasked with selecting the most likely correct ASR transcription from a list of candidates. As illustrated in Figure~\ref{fig:prompt}, all the N-best sentences are listed as input, such as 
\texttt{$<$option1$>$ ASR hypothesis $<$/option1$>$}, 
and ChatGPT is instructed to return the selected option in the format of 
\texttt{<option?> The selected ASR transcription </option?>}.
While this method is similar to language model rescoring to some extent, it differs in that the selection occurs in a single step. More importantly, ChatGPT sees all the candidates before determining the best one. This contrasts with the rescoring process, where language model scores are generated individually for each of the N-best hypotheses without considering their similarity and correlation.

\subsection{N-best Closest Decoding}
The closest mapping method \textbf{(closest)} is based on the assumption that during unconstrained error correction, LLMs first choose the best hypothesis from the given N-best list and then modify this sentence to yield the final output. In experiments, we aim to identify this \textit{``closest match''} by performing a reverse process, in which we locate the hypothesis within the ASR N-best list that has the smallest Levenshtein distance to the unconstrained generation result, as shown in Equation~\ref{eq:closest}:
\begin{equation}
\label{eq:closest}
\begin{split}
     \hat{\bm y}_\text{uncon} &= {\arg\max}_{\bm y}\log P(\bm y|\bm{\mathcal{Z}};\bm \theta_\text{EC}) \\
     \hat{\bm y}_\text{closest} &= {\arg\min}_{\bm z\in \mathcal{Z}}\text{LevenshteinDist}(\hat{\bm y}_\text{uncon}, \bm z) \\
\end{split}
\end{equation}
To illustrate, consider the \textit{zero-shot uncon} example in Figure \ref{fig:prompt}, where the Levenshtein distance of the ChatGPT output to the 3-best ASR hypotheses is $1, 0, 1$, respectively. In this scenario, the second hypothesis would be selected as the corrected result for the given utterance. 
Notably, this method is different from the N-best constrained method, which explicitly tasks the LLMs with selecting from the N-best list. The N-best closest approach does not constrain the output space initially but rather finds the closest match within the N-best list after the unconstrained generation.

\subsection{Lattice Constrained Decoding}
\begin{figure}[!t]
\centering
\subfloat[]{\includegraphics[width=0.36\textwidth]{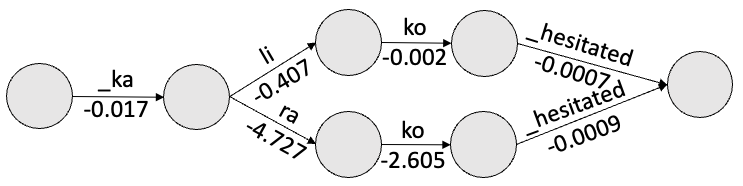}%
\label{fig:bpe}}
\vspace{-2mm}
\hfil
\subfloat[]{\includegraphics[width=0.26\textwidth]{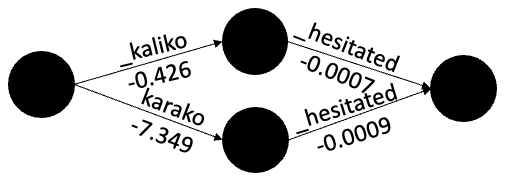}%
\label{fig:word}}
\vspace{-2mm}
\hfil
\subfloat[]{\includegraphics[width=0.48\textwidth]{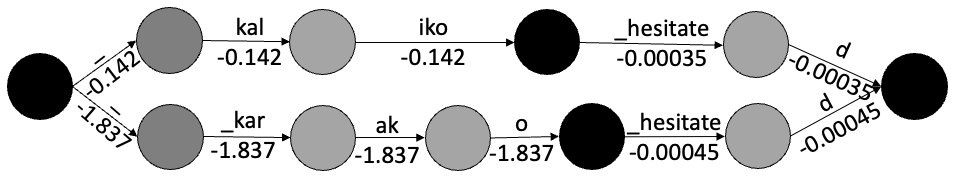}%
\label{fig:lm_bpe}}
\vspace{-2mm}
\caption{(a) Example BPE-level lattice generated in ASR decoding. (b) Converted word lattice. (c) Converted lattice with LLM BPE tokens.}
\label{fig:lattice}
\end{figure}

\begin{algorithm}[!htbp]
\footnotesize
\caption{Lattice Constrained Decoding for N-best T5}
\label{lattice_decode}
\textbf{Data:} lattice node set $\mathcal{V}$, lattice edge set $\mathcal{E}$, beam width $b$, T5 encoder outputs $\{\textbf h_j\}$

\begin{algorithmic}[1]
\STATE $Q\leftarrow \text{topological\_sort}(\mathcal{V})$
\FOR{$v$ in $\mathcal{V}$}
\STATE $H_v\leftarrow$ min\_heap()
\ENDFOR
\STATE $n_0.\text{history}=\epsilon, n_0.\text{score}=0$
\STATE $H_\text{start}$.put($n_0$) 
\FOR{$v$ in $Q$}
\STATE $\mathbf o=\text{Decoder}(\{\textbf h_j\}, n.\text{history}, v.\text{word}$)
\FOR{$\langle v,x\rangle$ in $\mathcal{E}$}
\FOR{$n$ in $H_v$}
\STATE $n^\prime.\text{history}=\text{concat}(n.\text{history},v.\text{word})$
\STATE $n^\prime.\text{score}=n.\text{score}+\lambda\cdot \log(\mathbf o[x.\text{word}])+(1-\lambda)\cdot\log s_{vx}$
\IF{$H_x.\text{size}\ge b\land H_x.\text{score}.\text{min}()<n^\prime.\text{score}$}
\STATE $H_x$.remove\_min()
\ENDIF
\IF{$H_x.\text{size}< b$}
\STATE $H_x$.put($n^\prime$)
\ENDIF
\ENDFOR
\ENDFOR
\ENDFOR
\STATE $I=$ max\_heap($H_\text{end}$.items)
\RETURN $I$.max()
\end{algorithmic}
\end{algorithm}

While an ASR N-best list is useful for capturing likely correct candidates, it represents only a limited set of possible decoding outputs. Instead of strictly constraining to the N-best list, we can explore a more flexible approach by expanding the decoding space to include the lattice generated through path merging. As depicted in Equation~\ref{eq:lattice}, we focus on the paths  $\bm{\mathcal{G}}$ within the lattice and integrate ASR scores during decoding. This lattice-constrained decoding approach enhances flexibility, allowing for a broader exploration of the potential corrections beyond the N-best list, and hence has the potential to improve overall decoding accuracy.
\begin{equation}
\begin{split}
    \hat{\bm y}_\text{lattice} &= {\arg\max}_{\bm y\in \bm{\mathcal{G}}} [(1-\lambda) \cdot \log P(\bm y|\bm{x};\bm\theta_\text{ASR})\\
    &+ \lambda \cdot \log P(\bm y|\bm{\mathcal{Z}};\bm \theta_\text{EC})]
\end{split}
\label{eq:lattice}
\end{equation}


Since this method requires ASR decoding probabilities at each time step to generate the lattice, it needs access to the ASR model, hence it is only applicable in certain scenarios. We tested this approach in the supervised EC model, N-best T5 model to be specific. 
Notably, the ASR model and the pre-trained language model employ different tokenizers. Consequently, we need to convert the original lattice into an equivalent form suitable for the N-best T5 to process. To achieve this, the lattice with ASR BPE tokens is first converted into a word lattice with dynamic programming, as shown in Figure \ref{fig:word}. Then the words along the edges are segmented into BPE tokens using the T5 tokenizer, as demonstrated in Figure \ref{fig:lm_bpe}. The decoding algorithm is adapted from \cite{auli2013joint}, and the details presented in Algorithm 1.
This method requires access to the beam search candidates in the decoding process, we only applied it to T5 rather than the closed-form LLM ChatGPT in the experiments. Our preliminary experiments indicate that performance remains consistent as beam sizes increase while the associated costs rise. Therefore in our experiments, we run the lattice-constrained decoding with the beam size of 1.

\section{Data Contamination}
\label{sec:contamination}
In the context of commercial use, most generative LLMs are developed with closed access to their training data. This lack of transparency presents challenges in determining if specific test data has been encountered during the model’s pre-training phase. Utilizing test data that has already been exposed to the LLM during pre-training can skew evaluations, resulting in inaccurate assessments of model performance -- an issue commonly referred to as data contamination \cite{sainz2023nlp, li2024task}. 

\begin{figure}
    \centering
    \includegraphics[width=1.0\linewidth]{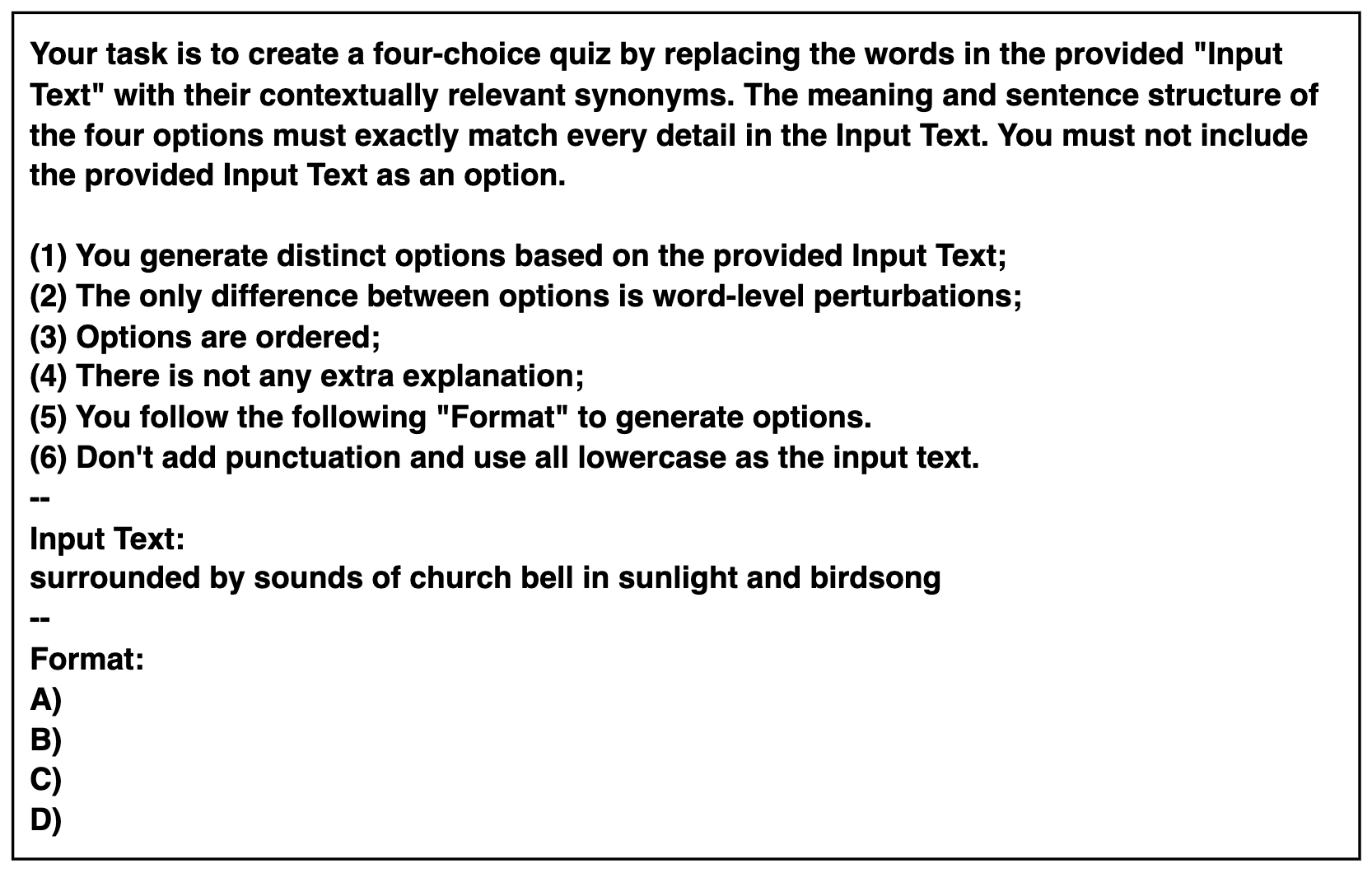}
    \caption{Prompt for generating options for the data contamination quiz.}
    \label{fig:paraphrase}
\end{figure}

To address this concern, we adapted the method from~\cite{golchin2023data} to measure the degree of the data contamination problem on the datasets we utilized in this paper. For each test utterance, we use GPT-4 to rewrite the original manual reference and generate paraphrased candidates. As presented in Figure \ref{fig:paraphrase}, instructions are given to GPT-4 to generate options by altering the words without affecting the sentence's meaning. We then randomly select one paraphrased sentence and feed it to LLM along with the original test sample to evaluate the degree of data contamination.

\begin{figure}
    \centering
    \includegraphics[width=1.0\linewidth]{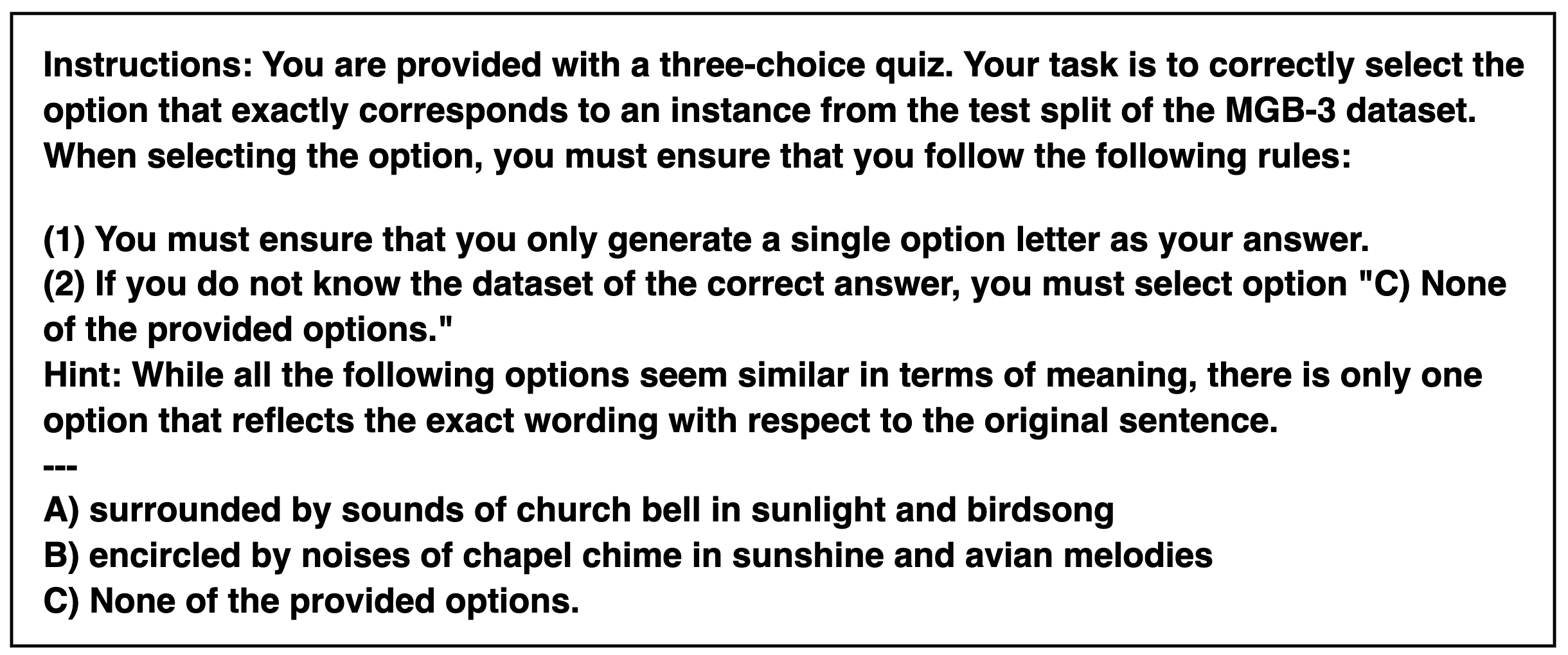}
    \caption{Demonstration of the data contamination quiz. Here, answer A) is selected from the original test set, and B) is the paraphrased one. We also switch the order of options A) and B) to mitigate the possible positional bias.}
    \label{fig:contam}
\end{figure}

Figure \ref{fig:contam} depicts the basic format of the designed 3-choice data contamination quiz. In this example, both sentences convey similar meanings while answer A) is copied from the test set and answer B) is the paraphrased one generated by a LLM. In addition, we provide option C), which denotes the non-appearance of both sentences. If the model generates A) in this scenario, it will suggest potential data contamination in the model pre-training. We estimate the level of contamination with the percentage of the test samples where the model selected the original sentence. 
This value is expected to be close to zero for a model that has never been pre-trained on the test set. Previous works have observed implicit positional bias in the LLM generation process \cite{liusie2023mitigating}. To target this problem, we run the method twice, changing the order of the given options, and compute the overall classification performance.

\section{Experiments}
\label{sec:exp}
\subsection{Experimental Setup}
Three standard datasets are used for training and evaluation, namely
LibriSpeech~\cite{panayotov2015librispeech}, TED-LIUM3~\cite{hernandez2018ted}, and Artie bias corpus~\cite{meyer2020artie}.
LibriSpeech is an audiobook-based English speech corpus, which covers a wide range of speakers with different accents.
TED-LIUM3 is an audio dataset collected from TED talks, encompassing various topics such as science, education, and entertainment. The Artie bias corpus is a subset of the Common Voice dataset~\cite{ardila2020common} which is also read speech. Detailed statistics of these datasets are listed in Table \ref{tab:statics}. 

\begin{table}[!t]
    \centering
    \caption{Statistics of ASR test sets used in the experiments.}
    \begin{tabular}{l|c|c|c|c}
    \toprule
        Dataset & Subset & \# Utts & \# Words & Hours \\
        \midrule
        \multirow{3}*{LibriSpeech}
        & train & 281,231 & 9.4M & 960.9 \\
        & test\_clean & 2,620 & 53K & 5.4 \\ 
        & test\_other & 2,939 & 52K & 5.1 \\ 
        \midrule
        TED-LIUM3 & test & 1,155 & 28K & 2.6 \\ 
        Artie Bias & test & 1,712 & 15K & 2.4 \\
    \bottomrule
    \end{tabular}
    \label{tab:statics}
\end{table}

The experiments were conducted on two ASR models: a Conformer-Transducer model~\cite{gulati2020conformer} and the OpenAI Whisper ASR~\cite{radford2023robust}. The Conformer-Transducer's encoder features 12 Conformer layers with a hidden size of 512 and its predictor has one LSTM layer. Both the jointer and predictor have hidden dimensions of 512. This model is trained on the 960hr LibriSpeech dataset following the ESPnet recipe \cite{watanabe2018espnet}. SpecAugment \cite{park2019specaugment} and speed perturbation are used for training data augmentation. We use a beam size of 10 in the decoding and save the generated top hypotheses. For Whisper, we adopt the small.en model due to its comparative performance to larger models and faster processing speed. The original decoding result only returns the 1-best hypothesis and the sentence-level confidence score for each utterance. We modified the code to also save the 10-best lists during inference and to extract token-level softmax probabilities for calculating word-level ASR confidence scores. This is to simulate the scenario where the ASR service provides extra information for downstream tasks. 
In the evaluation, we run text normalization scripts on both reference and hypothesis before calculating WER results following \cite{radford2023robust}.

For our fine-tuning ASR error correction method, we experimented with a T5 base model, an encoder-decoder model pre-trained on various text-to-text tasks. We aimed to build each EC model using the same ASR training corpus with input transcriptions generated by the corresponding trained ASR system. The Transducer model fit the ASR training set so well that it achieved an extremely low WER, making the development of an ASR error correction model impractical. To address this, we employed data augmentation methods to generate erroneous transcriptions for training the correction model. Specifically, we applied SpecAugment to each utterance in the ASR decoding process. 
We utilized two frequency masks with $F=30$, eight time masks with $T=40$, and time warping with $W=40$ on the training speech data. In the decoding results, sentences with WERs higher than 0.25 were filtered out, resulting in a training text corpus comprising 262K sentence pairs. The T5 model is fine-tuned for 3 epochs on this corpus using the AdamW~\cite{loshchilovdecoupled} optimizer. The initial learning rate is set to 5e-5 and the training batch size is 32. A dropout rate of 0.1 is applied to the network to prevent overfitting. For the zero-shot experiments, we used two versions of ChatGPT models: gpt-3.5-turbo-0613 and gpt-4-0125-preview, which are abbreviated into GPT-3.5 and GPT-4 in the paper.

\subsection{Experiments on Fine-tuning Approach}
\label{sec:exp-t5}
Table~\ref{tab:transducer_whisper_decoding} presents the results of two supervised error correction models trained for two ASR models with their outputs, evaluated on the LibriSpeech test\_clean and test\_other datasets.  For the Transducer model, the oracle WER improves by 33.5\% with the 5-best list and 38.4\% with the 10-best list on the test\_other set compared to the baseline. Similarly, for the Whisper model, the 5-best and 10-best lists achieve oracle WER improvements of 29.0\% and 35.8\% on the same set. These results suggest that the N-best lists have potential in helping the models recover the correct transcription.

\begin{table}[!t]
    \centering
    \caption{Results (\% WER) for a Conformer-Transducer system and Whisper using a T5 error correction model, comparing different (un)constrained decoding algorithms.}
    \begin{tabular}{l|l|cc|cc}
        \toprule
        \multicolumn{2}{l|}{\multirow{2}{*}{System}} & \multicolumn{2}{c|}{Transducer} & \multicolumn{2}{c}{Whisper} \\
        \multicolumn{2}{l|}{} & clean & other & clean & other \\
        \midrule
        \multicolumn{2}{l|}{Baseline} & 2.79 & 6.90 & 3.52 & 7.37 \\
        \multicolumn{2}{l|}{5-best Oracle} & 1.42 & 4.59 & 2.38 & 5.24 \\
        \multicolumn{2}{l|}{10-best Oracle} & 1.31 & 4.25 & 2.14 & 4.73 \\
        \midrule
        \multirow{4}{*}{\rotatebox{0}{10-best T5}} & uncon & 2.54 & 6.37 & \textbf{2.90} & \textbf{6.39} \\
        & constr & 2.42 & 6.15 & 3.10 & 6.69 \\
        & closest & 2.50 & 6.24 & 3.11 & 6.52 \\
        & lattice & \textbf{2.41} & \textbf{6.10} & - & - \\
        \midrule
    \end{tabular}
    \label{tab:transducer_whisper_decoding}
\end{table}

We compare the WER results of the 10-best T5 model using various decoding algorithms, as detailed in Section~\ref{sec:uncon_cons_decoding}. The EC model trained for the Transducer model with its outputs shows improved performance when additional constraints are applied during decoding. Specifically, in the constrained decoding process, optimal interpolation weight $\lambda$ is searched within the range [0.0, 1.0] with a grid size of 0.05. With N-best constrained decoding and closest mapping, the model effectively generates homophones for the mistaken words. Lattice-constrained decoding, which provides more potential paths than an N-best list and thus has a lower oracle WER, results in slightly better performance on the test sets (13.2\% and 11.6\% WERR on test\_clean and test\_other sets, respectively).
Unlike the N-best T5 for the Transducer ASR, the EC model tailored for Whisper outputs adeptly detects and corrects errors for LibriSpeech corpus utterances in unconstrained decoding, yielding WERR of 17.6\% on test\_clean and 13.3\% on test\_other. This indicates that the model has effectively learned to correct errors in the Whisper ASR from its training on 960hr LibriSpeech speech.
Constrained decoding yields less improvement. This limits the model to the N-best list which might have restricted performance compared to unconstrained decoding. 
Given the strong results from unconstrained decoding with Whisper and the complexity involved in generating lattices, lattice-constrained decoding is not considered here.
%


Results in Table~\ref{tab:transducer_whisper_decoding} have shown a significant improvement in recognition accuracy using the fine-tuned N-best T5 method on the target test sets. However, it will be more applicable if it can be applied to out-of-domain datasets or on outputs from a different ASR system. We therefore evaluate the generalization ability of the fine-tuning EC method from both perspectives.

\textbf{Generalization on out-of-domain datasets}: To examine the generalization ability of the proposed method on out-of-domain datasets, we directly applied the EC models trained on LibriSpeech transcriptions to the ASR outputs from other test sets (TED-LIUM and Artie bias corpus) without fine-tuning. The results are presented in Table~\ref{tab:supervised_ec_generalization}, and we studied the performance of two supervised EC models trained with Transducer ASR and Whisper ASR outputs, respectively.
In the baseline results, Whisper achieved much lower WERs for these datasets than the Transducer model, which was only trained with LibriSpeech data.  
The N-best T5 model, trained for Transducer outputs, improves ASR performance on out-of-domain datasets like TED and Artie without any fine-tuning, in both unconstrained and constrained decoding settings. Unconstrained decoding yields slightly better performance than N-best constrained decoding,  with relative word-error-rates (WERRs) of 11.3\% and 10.3\% on TED and Artie, respectively.
Although Whisper demonstrates very low baseline WERs, making further improvements challenging in a zero-shot transfer setting, the N-best T5 model still manages to reduce WER by 6.4\% and 9.9\% for TED and Artie, respectively, using N-best constrained decoding.

\textbf{Generalization on other ASR systems}: The practical utility of the EC model increases significantly if it can effectively correct outputs from ASR systems different from the one used for its training. In Table~\ref{tab:supervised_ec_generalization}, we investigated this aspect of the model's generalization ability. We applied the EC model trained with LibriSpeech transcriptions generated with the Whisper model directly to the ASR outputs from the Transducer model. Under unconstrained decoding conditions, the model struggled to achieve performance gains, highlighting the challenge of domain mismatch. However, employing constrained decoding successfully improved performance across both LibriSpeech test\_clean and test\_other datasets, resulting in a reduction of WER by 6.1\% and 3.6\%, respectively. This underscores the robustness of the proposed EC method in addressing different ASR system outputs.

\begin{table}[!t]
    \centering
    \caption{Error correction results (\% WER) for Conformer-Transducer and Whisper systems on out-of-domain ASR test sets and on outputs from a different ASR.}
    \begin{tabular}{@{ }l@{ }|@{ }l@{}|cc|cc|cc@{ }}
        \toprule
    \multicolumn{2}{l|}{EC training source} & \multicolumn{2}{c|}{Transducer} & \multicolumn{2}{c|}{Whisper} & \multicolumn{2}{c}{Whisper} \\
        \multicolumn{2}{l|}{EC applied to source} & \multicolumn{2}{c|}{Transducer} & \multicolumn{2}{c|}{Whisper} & \multicolumn{2}{c}{Transducer} \\
        \multicolumn{2}{l|}{Test sets} & TED & Artie & TED & Artie & clean & other \\
        \midrule
        \multicolumn{2}{l|}{Baseline} & 13.53 & 23.67 & 3.89 & 9.03 & 2.79 & 6.90\\
        \multicolumn{2}{l|}{10-best Oracle} & 10.21 & 16.69 & 2.59 & 5.60 & 1.31 & 4.25\\
        \midrule
        \multirow{2}{*}{\rotatebox{0}{10-best T5}} & uncon & \textbf{12.00} & \textbf{21.24} & 4.56 & 9.16 & 3.86 & 7.72\\
        & constr & 12.12 & 21.36 & \textbf{3.64} & \textbf{8.14} & \textbf{2.62} & \textbf{6.65}\\
        \midrule
    \end{tabular}
    \label{tab:supervised_ec_generalization}
\end{table}

\subsection{Experiments on Zero-shot Approach}
\label{sec:exp_zero-shot}
\noindent
This section will introduce zero-shot experiments and results using LLMs.
Table~\ref{tab:ec_big_table} presents the performance of ChatGPT (GPT-3.5 and GPT-4) for ASR error correction using either a Transducer-based ASR model or a Whisper model on the test\_other data set. We compare zero-shot error correction results under different decoding constraints. 
GPT-3.5 with unconstrained decoding shows improvement, however, further analysis reveals an increase in deletion errors on the test sets. This is due to the fact that some sentences are truncated in the ChatGPT output, displaying only the initial few words instead of complete sentences. Constrained generation methods limit the output to the N-best list, effectively reducing deletions. 
The \textit{closest} method, which identifies the closest hypothesis in the N-best list with the generated corrected one, outperforms methods that require ChatGPT to directly select the best option from the N-best list. Compared to GPT-3.5, GPT-4 shows improved performance on all test sets. The results indicate that for a powerful LLM like GPT-4, unconstrained decoding, giving the model more freedom, yields better results than constrained decoding approaches.


When comparing the zero-shot method with the fine-tuned N-best T5 model, we notice that GPT-3.5 matches the performance of the 10-best T5 for the Transducer ASR, while GPT-4 exceeds it across all three test sets. 
However, this success does not extend to outputs from the Whisper ASR, where LLMs in the zero-shot setting struggle more with error detection and correction. 
Using the closest match decoding approach, GPT-3.5 outperforms the baseline for the LB test set and Artie but does not reach the 10-best T5's performance. Notably, even GPT-4 falls short of surpassing the baseline on the TED test set. The ineffectiveness of zero-shot methods on TED-LIUM3 decoded by Whisper will be discussed in Section \ref{sec:discuss_n-best}). Specifically, GPT-4 achieves an average WERR of 25.2\% on the three test sets for Transducer outputs, while only an average of 2.6\% WERR is achieved on the three sets for Whisper outputs.
This is likely due to the nature of the N-best lists, Section~\ref{sec:discuss_n-best} will discuss this in detail.


\begin{table}[!t]
    \centering
    \caption{Results (\% WER) for a Conformer-Transducer system and Whisper with Zero-shot error correction using GPTs.}
    \begin{tabular}{@{ }l|l|ccc|ccc@{ }}
        \toprule
        \multicolumn{2}{@{ }l|}{\multirow{2}{*}{System}} & \multicolumn{3}{c|}{Transducer} & \multicolumn{3}{c}{Whisper} \\
        \multicolumn{2}{l|}{} & LB & TED & Artie & LB & TED & Artie \\
        \midrule
        \multicolumn{2}{@{ }l|}{Baseline} & 6.90 & 13.53 & 23.67 & 7.37 & \textbf{3.89} & 9.03 \\
        \multicolumn{2}{@{ }l|}{5-best Oracle} & 4.59 & 10.71 & 17.95 & 5.24 & 2.59 & 5.59 \\
        \multicolumn{2}{@{ }l|}{10-best T5} & 6.10 & - & - & 6.39 & - & - \\
        \midrule
        \multirow{3}{*}{\rotatebox{0}{GPT-3.5}} & uncon & 6.64 & 11.35 & 18.73 & 7.71 & 5.84 & 8.30 \\
        & constr & 6.52 & 12.61 & 21.88 & 7.24 & 4.19 & 8.47 \\
        & closest & 6.29 & 11.97 & 20.64 & 7.15 & 4.56 & 8.21 \\
        \midrule
        \multirow{3}{*}{\rotatebox{0}{GPT-4}} & uncon & \textbf{5.79} & \textbf{9.09} & \textbf{17.35} & \textbf{6.67} & 4.60 & \textbf{7.53} \\
        & constr & 6.55 & 11.91 & 20.97 & 7.17 & 4.58 & 8.46 \\
        & closest & 5.98 & 11.67 & 20.40 & 6.76 & 4.25 & 7.86 \\
        \bottomrule
    \end{tabular}
    \label{tab:ec_big_table}
\end{table}

\section{Discussion}
\label{sec:discussion}
\subsection{Ablation on N-best Inputs}
\label{sec:ablation_n}
This section presents the ablation experiments with diverse N-best inputs for fine-tuning and zero-shot EC methods.

\textbf{Fine-tuned EC approach}:
Table~\ref{tab:t5_ablation_n} presents results with baseline and N-best T5 models for the two ASR models on the LibriSpeech test\_clean and test\_other datasets. Training an EC model using the 1-best hypotheses from the Transducer ASR, as described in Section \ref{sec:ec_pre}, did not improve performance, highlighting the challenge of surpassing a strong baseline with limited input information. However, using 5-best and 10-best lists as model input, the T5 EC model can achieve relative performance gains of 6.0\% and 7.7\%  on the test\_other set, respectively. 
For the Whisper small.en model, the supervised error correction method showed performance gains even with the 1-best hypothesis, with improvements of 10.2\% on the test\_clean set and 4.6\% on the test\_other sets. Using the 5-best and 10-best lists yielded even better results, with the 10-best lists achieving a reduction of 17.6\% and 13.3\% on the test\_clean and test\_other sets, respectively. 
These findings indicate that while a larger N provides more diverse input and can improve error detection and correction, increasing N does not always guarantee proportional benefits.

\begin{table}[!t]
    \centering
    \footnotesize
    \caption{Results for ASR baseline and N-best T5 models with different model inputs on LibriSpeech test\_clean and test\_other sets.}
    \begin{tabular}{l|cc|cc}
    \toprule
    \multirow{2}*{Model} & \multicolumn{2}{c|}{Transducer} & \multicolumn{2}{c}{Whisper} \\
    & clean & other & clean & other \\
    \midrule
    Baseline & 2.79 & 6.90 & 3.52 & 7.37 \\
    \midrule
    1-best T5 & 2.94 & 7.00 & 3.16 & 7.03 \\
    5-best T5 & 2.63 & 6.49 & \textbf{2.86} & 6.59 \\
    10-best T5 & \textbf{2.54} & \textbf{6.37} & 2.90 & \textbf{6.39} \\
    \bottomrule
    \end{tabular}
    \label{tab:t5_ablation_n}
\end{table}

\begin{table}[!t]
    \centering
    \caption{Ablation of N-best list sizes utilizing GPT-3.5 for error correction on Conformer-Transducer outputs across three test sets with unconstrained decoding (\%WER).}
    \begin{tabular}{l|l|ccc}
        \toprule
        Method & Input & LB & TED & Artie \\
        \midrule
        Baseline & - & 6.90 & 13.53 & 23.67\\
        \midrule
        \multirow{4}*{GPT-3.5} & 1-best & 8.25  & 11.95 & 21.19 \\
        & 3-best & 7.01 & 11.31 & 18.84\\
        & 5-best & \textbf{6.64} & 11.35 & 18.73\\
        & 10-best & 6.69 & \textbf{11.29} & \textbf{18.72} \\
        \bottomrule
    \end{tabular}
    \label{tab:gpt_ablation_n}
\end{table}

\textbf{Zero-shot EC approach:} 
Table~\ref{tab:gpt_ablation_n} demonstrates the impact of varying N for the zero-shot error correction method, with the details of the method introduced in Section~\ref{sec:ec_zero-shot}. 
Specifically, we tested the GPT-3.5 model on the Transducer outputs across three datasets: LibriSpeech test\_other, TED-LIUM, and Artie, using the unconstrained decoding setup where the model generates sequences based on the input context.
The model faces challenges in enhancing performance when using only the top one ASR hypothesis as input. Notably, while performance improved on TED and Artie datasets, it declined on LibriSpeech. Increasing the number of input contexts generally helps the model detect and correct errors more effectively. However, our findings indicate that increasing N does not consistently improve results; for example, performance with the 10-best context was comparable to that with the 5-best. 
This finding is consistent with the observation on the fine-tuning EC method.

\subsection{Ablation on ASR Model Sizes}
\label{sec:ablation_whisper_size}
Previous experiments on Whisper are based on the small.en model, which yields good performance with relatively low decoding latency. In Table \ref{tab:ablation_whisper_size} we test Whisper models with different sizes as the underlying ASR model and apply GPT-4 as the correction approach. The results indicate that increasing the ASR model size generally improves baseline ASR performance, but also makes the correction task more challenging. Specifically, error correction yields higher WERR with smaller Whisper models. Except for Whisper large-v2, our method achieves a WERR ranging from 7.5\% to 17.6\%, demonstrating the effectiveness of zero-shot error correction.

\begin{table}[!t]
    \centering
    \caption{Ablation of zero-shot error correction results on different sizes of Whisper.}
    \renewcommand\tabcolsep{3.5pt}
    \begin{tabular}{@{ }l@{ }|c|cccc|cccc|c@{}}
    \toprule
    \multicolumn{1}{@{ }l|}{\multirow{2}{*}{System}} & Oracle & \multicolumn{4}{c|}{Baseline} & \multicolumn{4}{c|}{+GPT-4} & \multirow{2}{*}{WERR} \\
    \multicolumn{1}{l|}{ } & WER & All & Sub & Del & Ins & All & Sub & Del & Ins \\ 
        \midrule
        base.en & 6.75 & 9.50 & 7.1 & 1.0 & 1.4 & 7.91 & 5.6 & 1.0 & 1.3 & 17.6\% \\
        small.en & 5.24 & 7.37 & 4.9 & 1.7 & 0.8 & 6.67 & 4.3 & 1.6 & 0.8 & 9.5\% \\
        medium.en & 3.82 & 5.60 & 4.1 & 0.8 & 0.7 & 5.18 & 3.6 & 0.9 & 0.7 & 7.5\% \\
        large-v2 & 3.57 & 4.93 & 3.5 & 0.8 & 0.7 & 4.86 & 3.3 & 0.8 & 0.7 & 1.4\% \\
        \bottomrule
    \end{tabular}
    \label{tab:ablation_whisper_size}
\end{table}

\subsection{N-best Analysis}
\label{sec:discuss_n-best}
In Section~\ref{sec:ablation_n}, we observed that using an N-best list instead of the 1-best transcription significantly improves error correction performance. For the N-best T5 model, hypotheses are concatenated sequentially without explicitly encoding the ranking information. This raises an important question: Can the correction model infer and utilize this ranking knowledge from the input to enhance its performance? To investigate this, we conducted experiments with N-best lists that were either randomly shuffled or sorted in reverse order of ASR scores. As shown in Table~\ref{tab:whisper_t5_rescore_rnd}, applying unconstrained decoding to the LibriSpeech test sets and N-best constrained decoding to other datasets, we found that randomizing the N-best list led to performance degradation, while reversing the order of input hypotheses resulted in the worst performance. This indicates that the ranking information is implicitly learned and crucial for the N-best T5 model to perform well. However, when applying similar randomization or reversal strategies to the GPT-3.5 and GPT-4 models in zero-shot experiments, there was no significant difference in performance. This suggests that, unlike the N-best T5 model, the GPTs might not rely on or benefit from the ranking information in the same way.

\begin{table}[!t]
    \centering
    \caption{Ablation analysis with disturbed 10-best list for N-best T5 models.}
    \begin{tabular}{l|cc|ccc}
        \toprule
        \multirow{2}*{10-best} & \multicolumn{2}{c|}{LibriSpeech} & \multicolumn{3}{c}{Other sets} \\
        & clean & other & TED & Artie & MGB \\
        \midrule
        Sorted & \textbf{2.90} & \textbf{6.39} & \textbf{3.64} & \textbf{8.14} & \textbf{12.71} \\
        Randomized & 3.31 & 6.82 & 3.74 & 8.50 &  13.01 \\
        Reversed & 3.50 & 7.18 & 3.75 & 8.57 & 12.99\\
        \bottomrule
    \end{tabular}
    \label{tab:whisper_t5_rescore_rnd}
\end{table}

Experiments in Section~\ref{sec:exp_zero-shot} reveal that our proposed methods are less effective on Whisper outputs in some cases. To examine this, we calculate \textit{Uniq} and \textit{Cross WER} metrics in Table~\ref{tab:discuss_cross_wer_break}. 
When calculating statistics, we remove punctuation and special symbols from the ASR hypotheses, leaving only English characters and numbers to focus on the meaningful content.
The \textit{Uniq} metric represents the average number of unique hypotheses within an N-best list in the test set. For Transducer outputs, this number is close to 5, matching the size of the given N-best list. However, Whisper outputs show more repeated entries. This occurs because Whisper learns to generate sentences with inverse text normalisation (ITN) to enhance readability, i.e. adding capitalisation, including punctuation, and removing disfluencies. As a result, multiple hypotheses in an N-best list often differ only in format rather than content. This limits the diversity of the N-best list which is crucial for our proposed methods to work well.

Another notable observation is that for Whisper, even when the N-best list contains diverse hypotheses, the differences often come from the omission or insertion of irrelevant words.
This is demonstrated by the Cross WER metric in Table \ref{tab:discuss_cross_wer_break}. In this evaluation, we retain all unique hypotheses in an N-best list and then calculate the WER between each pair of hypotheses, summing the results for the entire set. This metric helps measure the difference between hypotheses within the same N-best list. The results indicate that Whisper has significantly higher deletion and insertion rates on Cross WER compared to the Transducer model, particularly on TED-LIUM3. This suggests that Whisper may struggle to consistently transcribe utterances accurately across all N-best hypotheses, resulting in sentences of varying lengths. ChatGPT tends to select the more coherent hypotheses in the zero-shot setting, leading to a higher rate of deletion errors in the output.

In Table \ref{tab:case_analysis} we demonstrate the outputs from different models on a specific example. Here, the unconstrained decoding is employed when 5-best lists are used as the model input for error correction. Compared to the desired reference, we highlight the errors in the generated ASR hypotheses and the error correction outputs from LLMs. The word \textit{ligatures}, a medical term related to surgery, is wrongly transcribed in the Top-1 ASR hypothesis. However, it appears in the correct format in the second-best hypothesis. With the T5 model, ASR errors are not recovered in the output. Meanwhile, as GPT models are pre-trained on more data, they present a better understanding of general world knowledge, leading to fewer errors utilizing the given contextual information.

\begin{table}[!t]
    \centering
    \caption{Statistics of the ASR 5-best lists generated by the Conformer-Transducer and the Whisper model on LibriSpeech (LB), TED-LIUM3 (TED) and Artie Bias test sets.}
    \begin{tabular}{l|l|c|c|c|c|c}
    \toprule
    \multirow{2}*{Data} & \multirow{2}*{Model} & \multirow{2}*{Uniq} & \multicolumn{4}{c}{\% Cross WER} \\
       &  &  & All & Sub & Del & Ins \\
      \midrule
        \multirow{2}{*}{LB} & Transducer & 4.9 & 9.1 & 7.1 & 1.0 & 1.0 \\
        & \multirow{1}{*}{Whisper} & 3.0 & 12.9 & 7.5 & 2.7 & 2.7 \\
        \midrule
        \multirow{2}{*}{TED} & Transducer & 5.0 & 7.4 & 5.4 & 1.0 & 1.0 \\
         & \multirow{1}{*}{Whisper} & 2.6 & 9.9 & 3.9 & 3.0 & 3.0 \\
         \midrule
        \multirow{2}{*}{Artie} & Transducer & 4.8 & 19.9 & 15.3 & 2.3 & 2.3 \\
         & \multirow{1}{*}{Whisper} & 2.9 & 21.1 & 14.5 & 3.3 & 3.3 \\
        \bottomrule
    \end{tabular}
    \label{tab:discuss_cross_wer_break}
\end{table}

\begin{table*}[h]
\centering
\caption{Case analysis for unconditional error correction results on Conformer-Transducer outputs.}
\footnotesize
\begin{tabular}{l|l}
\toprule
\multicolumn{1}{l|}{Ref} & the gut and the gullet being cut across between these ligatures the stomach may be removed entire without spilling its contents \\
\midrule
Hyp-1 & the gut and the gullet being cut across between these {\bf\color{red}ligatches} the stomach may be removed entire without {\bf\color{red}spinning} its contents \\
Hyp-2 & the gut and the gullet being cut across between these {\bf\color{blue}ligatures} the stomach may be removed entire without {\bf\color{red}spinning} its contents \\
Hyp-3 & the gut and the gullet being cut across between these {\bf\color{red}ligages} the stomach may be removed entire without {\bf\color{red}spinning} its contents \\
Hyp-4 & the gut and the gullet being cut across between these {\bf\color{red}ligatches} the stomach may be removed entire without {\bf\color{red}spinning as} contents \\
Hyp-5 & the gut and the gullet being cut across between these {\bf\color{red}ligatches} the stomach may be removed entire without {\bf\color{red}spinning his} contents  \\
\midrule
5-best T5 & the gut and the gullet being cut across between these {\bf\color{red}ligatches} the stomach may be removed entire without {\bf\color{red}spinning} its contents \\
GPT-3.5 & The gut and the gullet being cut across between these {\bf\color{blue}ligatures} the stomach may be removed entire without {\bf\color{red}spinning} its contents. \\
GPT-4 & The gut and the gullet being cut across between these {\bf\color{blue}ligatures} the stomach may be removed entire without {\bf\color{blue}spilling} its contents. \\
\bottomrule
    \end{tabular}
    \label{tab:case_analysis}
\end{table*}

\subsection{Multi-Model N-best Lists}

In previous experiments, we demonstrated that LLMs can make use of N-best lists generated by a single ASR model to perform error correction and enhance ASR performance. In this section, we extend this approach by combining the N-best decoding hypotheses from different ASR systems with LLMs. We experiment with two scenarios: (a) combining outputs from ASR models with different architectures and (b) combining outputs from ASR models trained on different datasets. 
ASR systems with different architectures exhibit unique strengths and weaknesses.
For instance, a LAS model is good at utilizing global context information from the given input but tends to be less robust. An RNN-T model alleviates the problem of repeating and skipping word chunks compared to a LAS model although it shows worse performance in general. By combining outputs, a more robust ASR system that takes advantage of different components can be built. Additionally, for ASR models trained on diverse datasets, combining the outputs acts as a form of model ensembling, thus is expected to improve the system performance. 

We also draw on the Recognizer Output Voting Error Reduction (ROVER) technique, which employs a majority voting approach to combine the recognition results of several ASR systems into a single recognition hypothesis~\cite{fiscus1997rover}. ROVER converts multiple ASR outputs into Word Transition Networks (WTNs), aligns and combines these WTNs using edit distance, and then uses weighted voting to determine the final hypothesis. Since ROVER is a simple, training-free technique for integrating information from different sentences, we use it as a baseline in the experiments.

\begin{table}[!t]
    \centering
     \caption{System combination results on test\_other using 5-best lists from Transducer ($T$), Whisper small.en ($E$) and small ($S$) models.}
    \begin{tabular}{l|l|ccc|cc}
    \toprule
        \multicolumn{2}{l|}{N-best Input} & T & E & S & Comb1 & Comb2  \\ 
        \midrule
        \multicolumn{2}{l|}{Baseline} & 6.90 & 7.37 & 7.20 & 5.95 & 6.82 \\
        \multicolumn{2}{l|}{5-best Oracle} & 4.59 & 5.24 & 5.21 & 3.35 & 4.40\\
        \midrule
        \multirow{2}*{GPT-3.5} & uncon & 6.64 & 7.71& 7.46 & 6.78 & 7.08 \\
        & closest & 6.29 & 7.15 & 7.06 & 6.01 & 6.46 \\
        \midrule
        \multirow{2}*{GPT-4} & uncon & 5.79 & 6.67 & 6.49 & \textbf{4.72} & 5.70 \\
        & closest & 5.98 & 6.76 & 6.51 & 5.00 & 5.80 \\
    \bottomrule
    \end{tabular}
    \label{tab:comb_model}
\end{table}

In Table \ref{tab:comb_model}, we list the ASR error correction results using the N-best list generated by a single system, denoted as $T_1T_2T_3T_4T_5$ (shortened as $T$) for Transducer outputs and $E_1E_2E_3E_4E_5$ (shortened as $E$) for Whisper small.en outputs.  $T_i$ and $E_i$ refer to the $i$-best hypothesis generated by the Transducer and the Whisper model respectively. For the output combination experiments, we take the 5-best lists generated by both the Transducer and the Whisper model for each test utterance. There are multiple ways to combine the two N-best lists to form a new 5-best list. In our preliminary experiments, we altered different ways of combining the N-best hypotheses and different sentence orders to find the best combination. We use ROVER to determine the performance of these different inputs, achieving the best performance with an input of $E_1E_2T_1T_2T_3$, denoted as $Comb1$ in Table \ref{tab:comb_model}, yielding a WER of 5.95\%.
Experiments on GPT models show that using outputs from diverse systems rather than a single system leads to performance boosts. LLMs could utilize information from both model outputs to generate a more robust answer. The best WER performance is 4.72\%, which is 32\% and 36\% lower than the Transducer ASR and Whisper baselines, respectively.

Additionally, we combine the ASR N-best lists generated by two different versions of Whisper models -- small and small.en, denoted as $S$ for N-best list $S_1S_2S_3S_4S_5$ from Whisper small model and $E$ for N-best list $E_1E_2E_3E_4E_5$ from Whisper small.en model, respectively. Both models are the same size but are trained on different training data. The small.en model was pre-trained on English-only weakly supervised data, while the small one was pre-trained on a larger, multilingual dataset. Experiments on ROVER show that the combination of $S_1S_2E_1E_2E_3$ ($Comb2$) works the best, leading to a WER of 6.82\% on the test set. 
The zero-shot error correction results indicate that LLMs serve as an effective method for model ensembling. With \textit{uncon} decoding using GPT-4, WERRs of 23\% and 21\% over ASR baselines can be seen on the test set. This showcases that LLMs can effectively improve ASR accuracy.

\subsection{Potential Data Contamination}
Previous results highlight the effectiveness of zero-shot error correction using LLMs, attributed to their robust language understanding capabilities. 
However, a key concern with this approach is the possibility that text from ASR test sets may have been included in the LLM pre-training data, potentially leading to biased evaluations in error correction tasks. In this section, we explore the potential issue of data leakage from ASR test sets during LLM pre-training.

Following the practice in \cite{golchin2023data}, we randomly select 100 utterances from each test set for evaluation. In addition to the three public test sets, we also applied our proposed method to two internal ASR datasets: MGB-3~\cite{bell2015mgb} and Linguaskill~\cite{xu2020linguaskill} that are less likely to be contaminated. These internal datasets are less susceptible to contamination due to their unique characteristics and restricted access. MGB-3 is a dataset specifically designed for the multi-genre broadcast challenge, containing broadcast media recordings that are carefully curated and controlled. Linguaskill, on the other hand, consists of educational and skill-based assessments that are not publicly available, ensuring a low risk of data contamination. Results on these datasets are intended to be contrasted with the results on the public datasets.

For the designed data contamination quiz, the LLM is asked to identify which sentence is from the ASR test set, choosing between an actual ASR test reference and a rewritten sentence. A lower percentage of correct selections by the LLM indicates a less severe data contamination.
The results in Table~\ref{tab:data_contamination} suggest that data contamination is not a significant issue for GPT-3.5. 
Although GPT-4 shows some level of data contamination on LibriSpeech and TED-LIUM3, this does not invalidate our method. The potential slight contamination observed with GPT-4 highlights an area for future improvement and caution but does not undermine the robustness and effectiveness of our proposed approach. The method shows consistent performance improvements across different datasets and ASR systems, indicating its generalizability and robustness despite the slight contamination in some instances.

\begin{table}[!t]
    \caption{Results of the data contamination quiz. A lower percentage implies less contamination in the LLM pre-training.}
    \centering
    \begin{tabular}{l|cc}
    \toprule
        Datasets & GPT-3.5 & GPT-4 \\
    \midrule
        LibriSpeech (test\_other) & 0.07 & 0.33 \\
        TED-LIUM 3 (test) & 0.05 & 0.22 \\
        Artie Bias (test) & 0.14 & 0.15 \\
        \midrule
        MGB-3 (test) & 0.04 & 0.02 \\
        Linguaskill (ling\_test\_general) & 0.03 & 0.09 \\
    \bottomrule
    \end{tabular}
    \label{tab:data_contamination}
\end{table}

\section{Conclusion}
\label{sec:conclusion}

In this work, we proposed and thoroughly investigated two advanced error correction methods to enhance ASR accuracy: supervised EC with pre-trained language models and zero-shot EC with LLMs. Various decoding strategies were explored for both supervised and zero-shot EC methods, including unconstrained decoding, N-best constrained decoding, and closest mapping decoding, each offering unique advantages in different scenarios. 
Our experiments demonstrated the robustness and generalization capabilities of the proposed methods across multiple dimensions. First, we tested our models trained on outputs from a specific ASR on outputs from different ASR systems, showcasing their adaptability. Second, we evaluated the models on datasets from diverse domains and applied an EC model trained on one dataset to other datasets, proving the versatility of the method. We also extended the approach to incorporate N-best lists from multiple ASR systems, demonstrating the model can serve as effective model ensembling.
Another crucial aspect of our study was addressing the potential data contamination issue, particularly in the use of LLMs for ASR error correction. Our systematic evaluation, using  a combination of public and proprietary datasets to ensure comprehensive coverage, revealed minimal contamination in GPT-3.5 but identified some level of contamination in GPT-4, especially on certain datasets. These findings underline the importance of vigilance in data handling and provide valuable guidelines for future research.

\section*{Acknowledgments}
This paper reports on research supported by EPSRC Project EP/V006223/1 (Multimodal Video Search by Examples) and Cambridge University Press \& Assessment, a department of The Chancellor, Masters, and Scholars of the University of Cambridge.

\newpage
\bibliographystyle{IEEEtran}
\bibliography{mybib}

\end{document}